\documentclass{article}

\PassOptionsToPackage{numbers, compress}{natbib}
\usepackage[preprint]{neurips_2023}

\usepackage[utf8]{inputenc} % allow utf-8 input
\usepackage[T1]{fontenc}    % use 8-bit T1 fonts
\usepackage{hyperref}       % hyperlinks
\usepackage{url}            % simple URL typesetting
\usepackage{booktabs}       % professional-quality tables
\usepackage{amsfonts}       % blackboard math symbols
\usepackage{nicefrac}       % compact symbols for 1/2, etc.
\usepackage{microtype}      % microtypography
\usepackage{xcolor}         % colors

\usepackage{multirow}
\usepackage{algorithm}
\usepackage{algorithmic}
\usepackage{subfigure}
\usepackage{multicol}

\usepackage{hyperref}
\usepackage{url}
\usepackage{comment}
\usepackage{graphicx}
\usepackage{amsmath}
\usepackage{color}
\usepackage{lipsum}
\usepackage{hanging}
\usepackage{enumitem}
\usepackage{booktabs}
\usepackage{wrapfig}

\title{GeneFace++: Generalized and Stable Real-Time Audio-Driven 3D Talking Face Generation}

\author{%
	Zhenhui Ye\thanks{Equal contribution.}   \thanks{Work done while at an internship at ByteDance.}   \\
	Zhejiang University\\
	\texttt{zhenhuiye@zju.edu.cn} \\
	\And
	Jinzheng He\footnotemark[1] \\
	Zhejiang University\\
	\texttt{jinzhenghe@zju.edu.cn} \\
	\And
	Ziyue Jiang\footnotemark[1] \\
	Zhejiang University\\
	\texttt{ziyuejiang@zju.edu.cn} \\
	\AND     
	Rongjie Huang \\
	Zhejiang University\\
	\texttt{rongjiehuang@zju.edu.cn} \\
	\And
	Jiawei Huang \\
	Zhejiang University\\
	\texttt{jiaweihuang@zju.edu.cn} \\
	\And
	Jinglin Liu \\
	Zhejiang University\\
	\texttt{jinglinliu@zju.edu.cn} \\
	\AND
	Yi Ren \\
	ByteDance \\
	\texttt{ren.yi@bytedance.com} \\
	\And
	Xiang Yin \\
	ByteDance \\
	\texttt{yinxiang.stephen@bytedance.com} \\
	\And
	Zejun Ma \\
	ByteDance \\
	\texttt{mazejun@bytedance.com} \\
	\And
	Zhou Zhao\thanks{Corresponding author} \\
	Zhejiang University \\
	\texttt{zhaozhou@zju.edu.cn} \\
}

\begin{document}

	\maketitle
	
%	
%	\begin{abstract}
%		Generating talking person portraits with arbitrary speech audio is a crucial problem in the field of the digital person and metaverse. Neural radiance field (NeRF)-based talking face generation systems such as GeneFace has achieved high-fidelity 3D-aware talking portraits synthesis with generalized audio-lip synchronization. However, there still exist several challenges for GeneFace: 1) as for the \textit{lip-synchronization}, it is hard to generate a long motion sequence of high temporal consistency with parallel computation; 2) as for the \textit{video quality}, the outliers of predicted facial motion occasionally produce bad rendering results; 3) as for the \textit{system efficiency}, the slow training and inference speed of the vanilla NeRF severely obstruct its usage in real-world applications. In this paper, we propose GeneFace++ to handle these challenges by 1) utilizing pitch contour as an auxiliary feature and enhancing the training loss of the audio-to-motion module to improve the time-consistency; 2) introducing locally linear embedding projection to regulate the outliers in the predicted motion sequence; 3) designing a computationally efficient NeRF-based motion-to-video renderer to achieves fast training and real-time inference. Extensive experiments show that our method achieves more generalized and high-fidelity talking face generation compared to previous methods. \footnote{Video samples are available at \url{https://genefaceplusplus.github.io}}
%	\end{abstract}

\begin{abstract}
Generating talking person portraits with arbitrary speech audio is a crucial problem in the field of digital human and metaverse. A modern talking face generation method is expected to achieve the goals of \textit{generalized audio-lip synchronization}, \textit{good video quality}, and \textit{high system efficiency}. Recently, neural radiance field (NeRF) has become a popular rendering technique in this field since it could achieve high-fidelity and 3D-consistent talking face generation with a few-minute-long training video. However, there still exist several challenges for NeRF-based methods: 1) as for the \textit{lip synchronization}, it is hard to generate a long facial motion sequence of high temporal consistency and audio-lip accuracy; 2) as for the \textit{video quality}, due to the limited data used to train the renderer, it is vulnerable to out-of-domain input condition and produce bad rendering results occasionally; 3) as for the \textit{system efficiency}, the slow training and inference speed of the vanilla NeRF severely obstruct its usage in real-world applications. In this paper, we propose GeneFace++ to handle these challenges by 1) utilizing the pitch contour as an auxiliary feature and introducing a temporal loss in the facial motion prediction process; 2) proposing a \textit{landmark locally linear embedding} method to regulate the outliers in the predicted motion sequence to avoid robustness issues; 3) designing a computationally efficient NeRF-based motion-to-video renderer to achieves fast training and real-time inference. With these settings, GeneFace++ becomes the first NeRF-based method that achieves stable and real-time talking face generation with generalized audio-lip synchronization. Extensive experiments show that our method outperforms state-of-the-art baselines in terms of subjective and objective evaluation. \footnote{Video samples are available at \url{https://genefaceplusplus.github.io}}
\end{abstract}

	\section{Introduction}
Audio-driven talking face generation is a popular topic in the field of the digital person and metaverse\cite{wav2lip, jamaludin2019you,ye2022syntaspeech}, aiming at synthesizing an audio-lip-synchronized video of the target person given the input audio. Among recent works, some of them generate video frames with convolutional neural networks trained with adversarial objectives\cite{Audio-driven_facial, LSP,suwajanakorn2017synthesizing} and suffer from an unstable training process and 3D in-consistency artifacts, while others \cite{guo2021ad,ye2023geneface} explore manipulating neural radiance field (NeRF)\cite{mildenhall2020nerf}, a more stable and 3D-aware model, to render talking face videos. In general, modern talking face generation systems aim to achieve the following goals:

\begin{itemize}[leftmargin=*]
	\item \textit{Generalized Audio-Lip Synchronization}: since people are sensitive to the slight misalignment between facial movements and speech audio \cite{chen2020comprises}, it is vital to maintain lip accuracy and temporal smoothness of the predicted facial motion. Considering the complex applications that may drive the system with out-of-domain (OOD) audio (such as cross-identity, cross-lingual, or singing audio), the model should also generalize well to various audios \cite{ye2023geneface}.
	\item \textit{Good Video Quality}: the overall good video quality typically consists of high image fidelity, smooth transition between adjacent frames, and realistic 3D modeling of the talking avatar.
	\item \textit{High System Efficiency}: to reduce the cost of computational resources and apply the model to real-world applications, the training process should be easy and the inference speed should be fast.
\end{itemize}

With the development of neural rendering techniques such as NeRF \cite{mildenhall2020nerf} and its variants \cite{EG3D}, it is possible to build a high-fidelity 3D-aware talking face generation system with a several-minute-long video of the target person. Hence the second objective, i.e., \textit{good video quality} has been partially achieved. However, as early NeRF-based methods \cite{guo2021ad}\cite{tang2022real} mainly train the model in an end-to-end manner, considering the limited amount of audio-lip pairs utlized in this process,
%\footnote{the end-to-end audio-to-video works have implicitly modeled the audio-to-motion mapping in the radiance field.}
 it is hard for them to achieve the goal of \textit{generalized audio-lip synchronization}. Recently, GeneFace\cite{ye2023geneface} partially handles this problem by introducing a generalized audio-to-motion mapping learned from large-scale lip-reading datasets to predict accurate motion representations for the motion-conditioned NeRF-based renderer. Despite the community's efforts to improve the practicality of NeRF-based talking face systems, there still exist several challenges that impede its usage in real-world applications:
\begin{itemize}[leftmargin=*]
	\item As for the \textit{audio-lip synchronization}, it is still challenging to model the long audio sequence and generate time-consistent results. For instance, when a drawl or trill in the singing voice occurs, a single phoneme may last for more than 2 seconds (about 50 video frames), which requires the predicted lip motion to be consistent in the long term to seem perceptually natural.
	\item As for the \textit{video quality}, it is hard to render diverse facial motions (such as an extra-big mouth) for a neural renderer that is only trained with consecutive frames from a few-minute-long video. GeneFace first notices this problem and learns a domain adaptative (DA) Postnet with adversarial training to map the predicted facial motion into the narrow input space of the renderer. However, it cannot guarantee the successful transformation of each frame and bad cases still occurs occasionally. 
	\item As for the \textit{system efficiency}, the slow training and inference speed caused by the expensive computation cost of vanilla NeRF has severely obstructed the usage in real-time applications.
\end{itemize}

In this work, we propose \textbf{GeneFace++} to address the three challenges in NeRF-based methods and achieve the goals of modern talking face generation. Specifically, 1) to improve long-term temporal consistency and naturalness of the predicted facial landmark sequence, we propose a \textit{Pitch-Aware Audio-to-Motion} module. Specifically, we introduce the pitch contour as an auxiliary feature of the audio-to-motion mapping. We show that pitch could act as a helpful facial motion indicator to improve lip-sync quality. We also introduce a temporal smoothing loss to improve the overall temporal stability of the generated landmark sequence. 2) To improve the robustness of the system to diverse facial motions, we propose a manifold projection-based method named \textit{Landmark Locally Linear Embedding} to post-process the predicted landmark, which solves a least square equation to reconstruct the input landmark with a linear combination of several nearest ground truth (GT) data points. It could map the landmark closer to the GT target person dataset and thus significantly reduce bad cases in rendering. 3) To improve the efficiency of the renderer, we propose an efficient dynamic NeRF named \textit{Instant Motion-to-Video} module,  which utilizes grid-based embedders to ease the training process and adopts deformable slicing surfaces to model the dynamics of the facial geometry conditioned on the 3D facial landmark. Compared with vanilla NeRF, the proposed renderer could be trained more efficiently and infer in real-time.

To summarize, GeneFace++ mitigates the difficulties in NeRF-based talking face video synthesis and achieves the goals of \textit{generalized audio-lip synchronization}, \textit{good video quality}, and \textit{high system efficiency}. GeneFace++ could transform various OOD voices into accurate and time-consistent facial landmarks, and render high-fidelity 3D-aware human portraits in a stable and efficient manner. Extensive experiments show that GeneFace++ outperforms other state-of-the-art audio-driven talking face generation methods from the perspective of objective and subjective metrics. Ablation studies prove the necessity of each component.

%The contributions can be summarized as follows:
%
%\begin{itemize}[leftmargin=*]
%	\item We propose a pitch-aware audio-to-motion module to model the long audio sequence and generate time-consistent results.
%	\item We present a LLE-based postprocessing strategy to refine the predicted facial motion and prevent rendering bad cases caused by outliers. We also design a computationally efficient NeRF-based renderer that could generate high-fidelity images in real-time.
%	\item Extensive experiments show that GeneFace++ outperforms other state-of-the-art audio-driven talking face generation methods from the perspective of objective and subjective metrics. Ablation studies prove the necessity of each component.
%\end{itemize}

\section{Related Works}
\label{app:related}
%Talking face generation aims at synthesizing talking person portraits in line with the driving audio, which can be divided into two sequential steps: an audio-to-motion process that predicts facial motion given the input audio, and a motion-to-video process that renders the human portrait image given the input facial motion. GeneFace++ is a 3D talking face system that aims to provide a more accurate and generalized audio-to-motion transform, as well as a robust and efficient motion-to-video renderer that could generate high-fidelity video frames. Therefore, we discuss related works about the lip-synchronized audio-to-motion transform and the human portrait rendering techniques, respectively. 
Talking face generation can be divided into two sequential steps: an audio-to-motion process that predicts facial motion given the input audio, and a motion-to-video process that renders the human portrait image given the input facial motion. We discuss related works about the lip-synchronized audio-to-motion transform and the human portrait rendering techniques, respectively.

\subsection{Lip-Synchronized Audio-to-Motion}

To achieve lip-synchronized motion prediction, there are two challenges faced by the community. The first challenge is the so-called one-to-many mapping problem, which means the same input audio may have several reasonable corresponding facial motions. Some early work \cite{zhou2020makelttalk, zhang2021facial, ATVG} directly learn a deterministic model with a regression loss (e.g., L2 error), and suffer from over-smoothed lip results. Wav2Lip \cite{wav2lip} first utilizes a discriminative sync expert to achieve a more sharp and accurate lip motion, which is followed by latter works\cite{zhou2019talking,yin2022styleheat,liang2022expressive, ji2022eamm, sun2021speech2talking}. MemFace\cite{tang2022memories} introduces memory retrieval in audio-to-motion to alleviate the one-to-many problem. The second challenge is to generate time consistent and stablized motion sequence given the long input audio. Some work \cite{LSP} adopt auto-regressive structure to model the temporal sequence, but suffer from slow inference and error accumulation. Other works \cite{styleavatar, guo2021ad} use parallel structure (such as 1D Convolution) with a sliding window, which partially address the shortage of auto-regressive methods. Recently, several works such as \cite{chen2022transformer} and GeneFace\cite{ye2023geneface} use feedforward structures (self-attention and convolution) to process the whole audio sequence in parallel. This framework enjoys high efficiency and capability to model the long term information, but is challenged in keeping temporal consistency and stability in the generated motion sequence.

\subsection{Human Portrait Rendering} 
The modern techniques for dynamic human portrait synthesis could be categorized into three classes: 1) 2D-based, 2) 3D Morphable Model \cite{paysan20093d}  (3DMM)-based,  and 3) neural rendering-based. The earliest works typically belong to the 2D-based methods\cite{thies2020neural,suwajanakorn2017synthesizing,wav2lip,zhou2020makelttalk,yu2020multimodal,zhou2019talking}, which either adopt GANs \cite{goodfellow2020generative} or image-to-image translation \cite{isola2017image} as the image renderer. Altough achieves good image fidelity, these methods fail to generate pose-controllable videos due a lack of 3D geometry modeling. The 3DMM-based methods\cite{styleavatar,Audio-driven_facial,yi2020audio} inject the 3D prior knowledge by using 3DMM coefficients as auxiliary conditions, but using 3DMM as the intermediate is known to cause an information loss, which degrades the performance. Recently, the neural rendering-based methods\cite{chan2021pi, NeRFACE, pumarola2021d, HeadNeRF, zhuang2022mofanerf} adopt NeRF\cite{mildenhall2020nerf} or its variants\cite{EG3D} to enjoy a realistic 3D modeling of the human head. AD-NeRF\cite{guo2021ad} is the first NeRF-based talking face method, which presents an end-to-end audio-to-video NeRF renderer to generate face images conditioned on audio features. Then some works explore to improve the sample efficiency \cite{liu2022semantic} and achieve few-shot synthesis \cite{shen2022learning}. To achieve good lip synchronization, GeneFace\cite{ye2023geneface} introduces an independent audio-to-motion module for the NeRF-based renderer. To improve the system efficiency, RAD-NeRF \cite{tang2022real} introduces discrete learnable grids \cite{instant-ngp} in AD-NeRF, which accelerates training and inference.

As discussed above, with the rapid advances in the audio-to-motion and motion-to-video process, a modern talking face system that simultaneously achieves the goals of "generalized audio-lip synchronization", "good video quality" and "high system efficiency" is dawning. This observation motivates the design of GeneFace++. As for the audio-to-motion phase,  we solve the long-term consistency problem with pitch information and a temporal smoothing loss. A manifold projection-based postprocessing method is also proposed to improve the robustness of the system. As for the motion-to-video phase, we utilize grid encoders and deformable slicing surfaces to achieve a high-quality and efficient motion-conditioned human portrait renderer.

\section{Preliminaries: GeneFace}
\label{app:geneface}
Our proposed GeneFace++ follows the two-stage paradigm in GeneFace \cite{ye2023geneface}. Therefore, in this section, we introduce preliminary knowledge about the \textit{audio-to-motion} and \textit{motion-to-video} stage in GeneFace.

\noindent \textbf{Audio-to-Motion}  At the \textit{audio-to-motion} stage, GeneFace first learns a conditional variational auto-encoder (VAE) \cite{kingma2013auto} model in a large-scale lip-reading dataset to achieve generalized and accurate facial landmark prediction given various audios. Specifically, the training loss of the VAE is:
\begin{equation}
	\mathcal{L}_{\text{VAE}}=\mathbb{E}[ ||\mathbf{l} - \hat{\mathbf{l}}||_2^2  + {KL}(z| \hat{z}) + l_{Sync}(\textbf{a}, \hat{\mathbf{l}}) ] , ~~ s.t. ~~~~ z\sim N(0,1), 
	\label{eq:vae}
\end{equation}
where $\textbf{a}$ is the input audio and $\mathbf{l}$ is the corresponding GT facial landamrk. $Enc$ and $Dec$ denote the encoder and decoder in VAE, respectively. $\hat{z}={Enc}(\mathbf{l}, \textbf{a})$ is the latent encoding of GT landmark and $\hat{\mathbf{l}}={Dec}(\hat{z}, \textbf{a})$ is the predicted facial landmark. $KL$ denotes KL divergence and $l_{Sync}$ is a perceptual loss that measures the audio-visual synchronization, which is provided by a pretrained sync expert \cite{wav2lip}. After training, the encoder is discarded and only the decoder is needed to predict the facial landmark given the input audio.

To overcome the significant domain gap between the large-scale lip-reading dataset and the target person video, GeneFace adopts adversarial domain adaptation to learn a domain adaptative (DA) Postnet that projects the predicted facial motion into the target person domain. The training loss of DA Postnet is:
\begin{equation}
	\mathcal{L}_{\text{Postnet}}=\mathbb{E}[ l_{Adv}(\overline{\mathbf{l}})) + l_{Sync}(\textbf{a}, \overline{\mathbf{l}})] , ~~s.t.~~\overline{\mathbf{l}}=PN(\hat{\mathbf{l}}),
	\label{eq:pn}
\end{equation}
where $\overline{\mathbf{l}}$ represents the refined landmark generated by the DA Postnet and ${PN}$ is the Postnet  (a shallow 1D convolutional neural network) to be trained. $\hat{\mathbf{l}}={Dec}(z, \textbf{a})$ is the facial landmark generated by the VAE decoder and $z$ is the noise sampled from normalized gaussian distribution. $ l_{Adv}$ is the LSGAN-styled\cite{mao2017least} adversarial loss, whose objective is to minimize the distance between the refined landmark distribution and the GT target person landmark set.

Once the training of the DA Postnet is done, the audio-to-motion module, which consists of a VAE decoder followed by a DA Postnet, could generate lip-sync and personalized facial landmarks given the input audio:
\begin{equation}
	\overline{\mathbf{l}}=PN(\hat{\mathbf{l}}) = PN(Dec(z, \textbf{a})).
\end{equation}

\noindent \textbf{Motion-to-Video}  As for the \textit{motion-to-video} stage, GeneFace designs a landmark-conditioned dynamic NeRF network to render the human portrait given the input facial landmark. Specifically, it learns an implicit function $F$ which can be formulated as follows:
\begin{equation}
	F: (x,d,\mathbf{l})\rightarrow (c, \sigma),
\end{equation}
where $x$ and $d$ are positions and view direction. $\mathbf{l}$ is the facial landmark that morphs the human head, which is implicitly modeled in the function $F$. $c$ and $\sigma$ denote the predicted color and density in the 3D radiance field. We can conveniently render an image from this radiance field via a differentiable volume rendering equation that aggregates the color $c$ along the ray $r$:
\begin{equation}
	C(r, \mathbf{l})=\int_{t_n}^{t_f} \sigma(r(t),\mathbf{l})\cdot c(r(t),\mathbf{l},d)\cdot T(t)dt ,
	\label{eq:render_image}
\end{equation}
where $C$ is the RGB value of the pixel that corresponds to the ray $r$ emitted in the 3D space. $t_n$ and $t_r$ are the near bound and far bound of the ray. $r(t)$ is a shorthand of the position $x$ and direction $d$ at the sampled point $t$ of the ray. $T(t)=\exp(-\int_{t_n}^{t}\sigma(r(\tau), \mathbf{l})d\tau)$ is the accumulated transmittance along the ray from $t_n$ to $t$. Now that the image could be rendered, the training objective of NeRF is to reduce the L2 error between the rendered and ground-truth images:
\begin{equation}
	\mathcal{L}_{\text{NeRF}}=\mathbb{E}[ ||C(r,\mathbf{l})-C_g||^2_2 ].
	\label{eq:train_nerf}
\end{equation}
During inference, by cascading through the \textit{audio-to-motion} module and the \textit{motion-to-video} module, GeneFace achieves to generate lip-sync and high-fidelity talking face video to various OOD audios.

\section{GeneFace++}
\begin{figure*}[ht]
	\centering
	\includegraphics[width=0.99\textwidth]{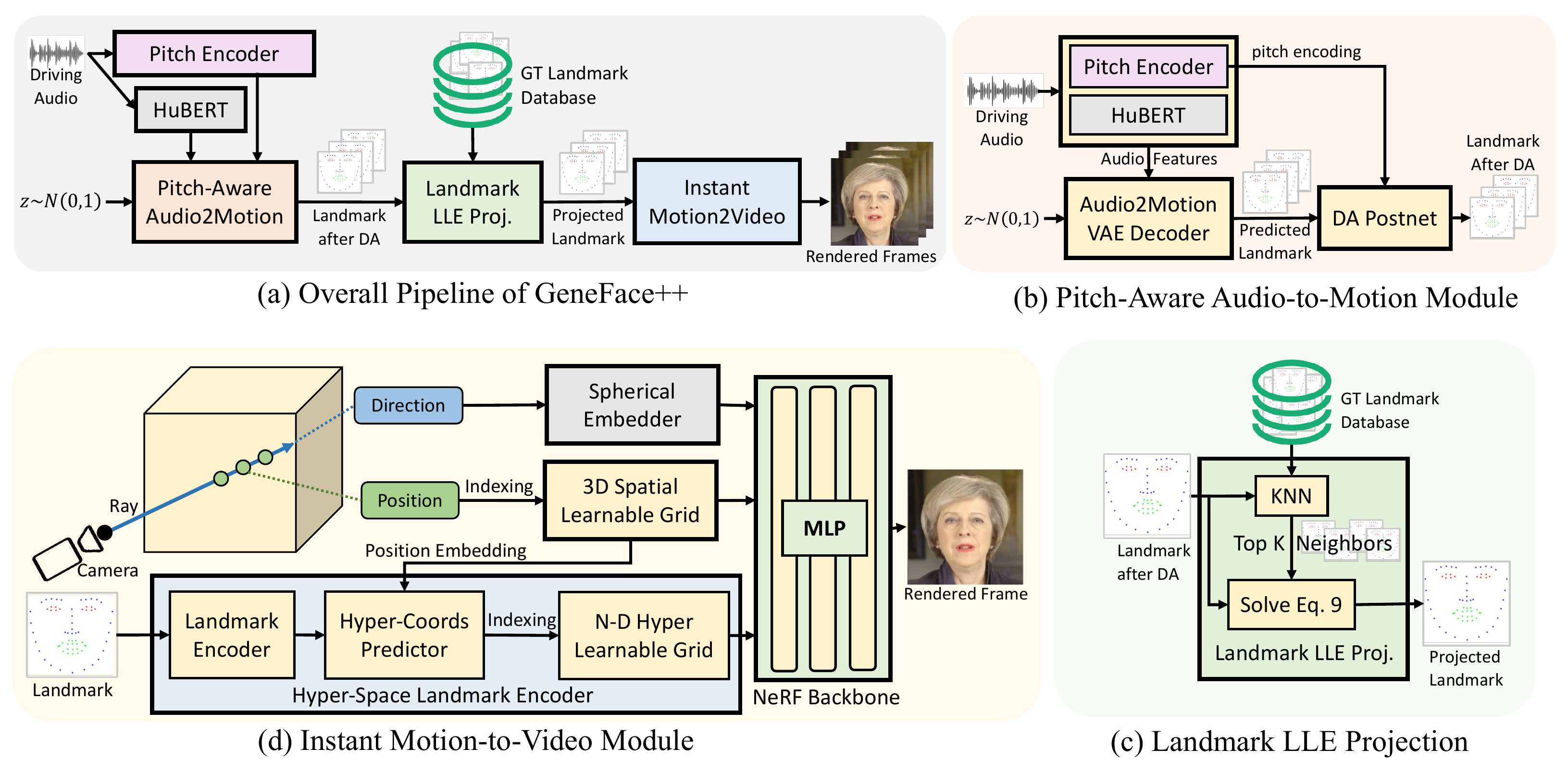}
	\caption{The inference process of GeneFace++. In subfigure (a), we show the overall three-stage pipeline. 
		In subfigure (b), "DA Postnet" denotes the Domain Adaptative Postnet proposed in GeneFace. In subfigure (c), "KNN" denotes finding the K-nearest neighbors of the input landmark, and "Landmark LLE Proj." denotes Landmark Locally Linear Embedding Projection method proposed in Section \ref{sec:lle}. In subfigure (d), as for the \textit{indexing} operation, we perform bi-linear interpolation to query the continuous coordinates in the discrete (spatial/hyper) grids.
	}
	\label{fig: overall}
\end{figure*}

In this section, we introduce the architecture of GeneFace++, which aims to improve GeneFace to achieve more natural \textit{audio-lip synchronization}, more robust \textit{video quality}, and higher \textit{system efficiency}. As shown in Fig. \ref{fig: overall}(a), GeneFace++ is composed of three parts: 1) a \textit{pitch-aware audio-to-motion} module that transforms audio features into facial motion; 2) a \textit{landmark locally linear embedding} method to post-process the predicted motion; and 3) an \textit{instant motion-to-video} module that could render the final talking face video efficiently. We describe the designs and the training process of these three parts in detail in the following subsections. We provide detailed network structures in Appendix \ref{app:structure}.
\subsection{Pitch-Aware Audio-to-Motion Transform}

The motivation for considering pitch information in the audio-to-motion mapping is that pitch is known to highly correlated to facial expressions \cite{thompson2010facial}. For instance, a high and steady pitch contour may correlate to a large and steady lip motion. Then we further found two advantages of using pitch in the audio-to-motion module: (1) We notice that previous NeRF-based methods \cite{guo2021ad}\cite{ye2023geneface} utilize phonetic posteriorgrams features \cite{hannun2014deepspeech}\cite{hsu2021hubert} as the audio feature to ease the training and improve cross-lingual cross-identity generalizability. However, since PPGs is intended for ASR usage, it ignores the acoustic information in the waveform, which is necessary to achieve stable and expressive facial motion prediction. Under this circumstance, auxiliary acoustic features such as pitch contours are helpful to improve the expressiveness and temporary consistency of the predicted facial motion. 
%\footnote{Pitch contour is a 1-dimensional acoustic signal, which could ease the model to perceive the high-level status (such as voiced/unvoiced) of the long audio sequence.}
(2) The second reason to introduce pitch information is the observation of the unstable performance of the DA Postnet in GeneFace: the Postnet is only provided with the predicted facial motion and is requested to project it into the target domain. It is hard for the Postnet to model this domain shift in its implicit space without any condition, which leads to unstable training %\footnote{For instance, mode collapse occurs or the Postnet is defeated by the discriminator} 
and occasional bad cases. We suggest that the pitch information could act as a lightweight and helpful hint for the Postnet to better process the input facial landmark.

\begin{figure*}[ht]
	\centering
	\includegraphics[width=0.99\textwidth]{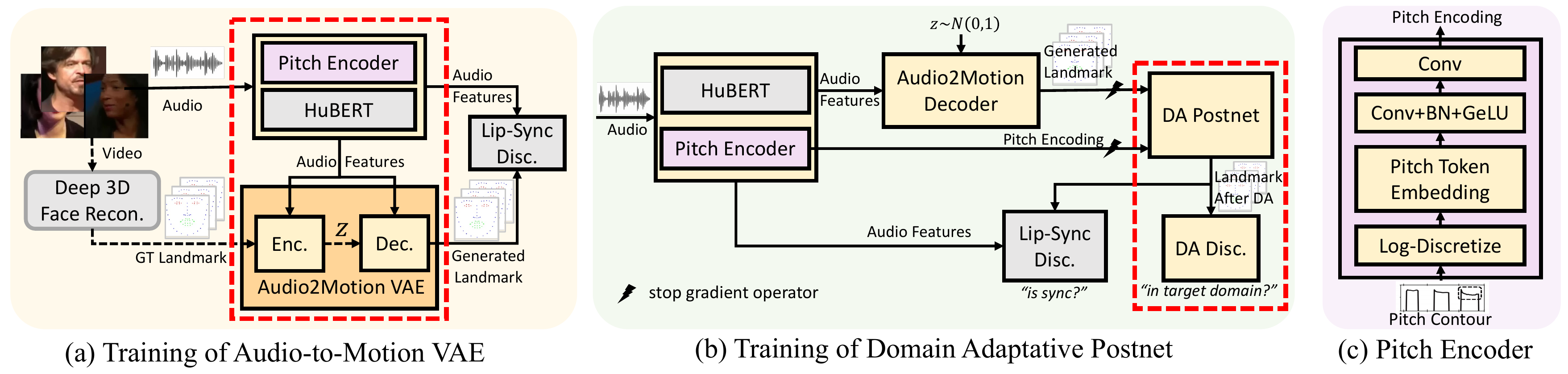}
	\caption{The training process of the Pitch-Aware Audio-to-Motion module. Learnable models are marked with red dotted rectangles and parameter-frozen models are colored in gray. In subfigure (a), Enc. and Dec. denote Encoder and Decoder in VAE, respectively. Disc. means Discriminator. In subfigure (b), the thunder-like symbol represents the "stop gradient" operator. In subfigure(c), "Log-Discretize" denotes the operation that quantizes the log-scale continuous pitch value into discrete tokens.
	}
	\label{fig: audio2motion}
\end{figure*}

\noindent \textbf{Pitch Encoder} 
As shown in Fig. \ref{fig: audio2motion}(c), we design a pitch encoder to effectively extract information from the pitch encoder. The key idea to designing the pitch encoder is to keep it lightweight and efficient. To be specific, we first discretize the continuous pitch (fundamental frequency) value into several discrete tokens to ensure temporary smoothness and ease the training of the pitch encoder. Note that we discretize the bins in log-scale to adhere to human perception. A group of pitch embedding corresponding to the discrete pitch tokens is learned from scratch during the training of the \textit{pitch-aware VAE} (discussed latter). The pitch embedding is fed into a shallow convolutional network, which consists of a 1D convolutional layer with batch normalization and GeLU activation, and an extra 1D convolutional layer to produce the final pitch encoding.

\noindent \textbf{Pitch-Aware VAE} 
As shown in  Fig. \ref{fig: audio2motion}(a), we plug the pitch encoder as an auxiliary condition encoder into the audio-to-motion VAE. We train the pitch encoder and VAE on a large-scale lip-reading dataset following Equation \ref{eq:vae}. During inference, the pitch encoder and VAE decoder are used to predict the facial landmark:
\begin{equation}
	\hat{\mathbf{l}} = Dec(z, \textbf{h}, PE(\textbf{p})) , ~~ s.t. ~~~~ z\sim N(0,1), 
	\label{eq:pitch_dec}
\end{equation}
where $\textbf{h}$ and $\textbf{p}$ are the HuBERT feature and pitch value extracted from the input audio. $PE$ represents the pitch encoder.

\noindent \textbf{Pitch-Aware DA Postnet} 
As shown in Fig. \ref{fig: audio2motion}(b), the pretrained pitch encoder provides the auxiliary pitch encoding to the DA Postnet. We train the DA Postnet following Equation \ref{eq:pn}. Since the supervision signal of the adversarial training process is known to be unstable, we fix the parameters of the pitch encoder and VAE decoder to prevent deteriorated performance. During inference, as shown in Figure \ref{fig: overall}(a), the pitch-aware decoder and DA Postnet could generate personalized facial landmark $\overline{l}$ given the input audio:
\begin{equation}
	\overline{\mathbf{l}} = PN(\hat{\mathbf{l}}, PE(\textbf{p})),
\end{equation}
where $\hat{\mathbf{l}}$ is the facial landmark predicted by decoder in Equation \ref{eq:pitch_dec}.

\noindent \textbf{Training} 
We follow the major setting of GeneFace to train the VAE  and Postnet with two modifications: (1)To remove the jitter in the predicted landmark, inspired by FACIAL\cite{zhang2021facial}, we encourage the VAE to optimize the velocity of the landmark sequence. Specifically, a temporal smoothing term is added to the training objective of VAE. (2) To stabilize the adversarial training, we adopt the gradient penalty in WGAN-GP\cite{gulrajani2017improved} when updating the discriminator of Postnet. Due to spatial limitation, we illustrate the modified training loss in Appendix \ref{app:loss_audio2motion}.

\subsection{Landmark Locally Linear Embedding}
\label{sec:lle}
By introducing pitch information into the audio-to-motion module, we have improved the temporal consistency and naturalness of the predicted landmark. However, improving the quality of predicted landmarks is not adequate to achieve \textit{good video quality}, as it requires the NeRF-based motion-to-video module to accurately render the human portrait corresponding to the assigned facial motion. This NeRF-based renderer, however, is typically learned from a very small dataset (a few-minute-long video), hence are expected to only work well on a narrow input space of facial landmark. When confronted with OOD landmarks, the renderer may produce inaccurate facial motion or even crashed rendering results. GeneFace utilize adversarial domain adaptation to train a Postnet to map all landmarks into the NeRF's narrow input space. However, due to the instability of adversarial training, it is not theoretically guaranteed to correctly project every frame into the target domain, and bad case occurs occasionally, which raises robust challenges to real-world applications.

In this context, as shown in Figure \ref{fig: overall}(c), we propose \textit{landmark locally linear embedding} (Landmark LLE), a manifold projection-based post-processing method that guarantees each predicted landmark is successfully mapped into (the vicinity of) the input space of the landmark-conditioned renderer. In other words, with the help of Landmark LLE, each predicted landmark is dragged closer to the GT landmark set that is used to train the renderer. 
%The motivation of Landmark LLE is inspired by recent work in face synthesis from sketches \cite{chen2020deepfacedrawing}, which adopts manifold projection to improve the robustness of OOD sketches.
 We follow the main idea of the classic locally linear embedding (LLE) algorithm \cite{LLE} on the facial representation manifold: each facial landmark and its neighbors are locally-linear on the manifold. Motivated by the success of 3DMM, which could reconstruct arbitrary human face with a linear combination of around 144 template meshes, we assume that the facial landmark data points themselves are locally-linear, and there is no need to project them into a higher dimension manifold as vanilla LLE does. To be specific, given a predicted 3D facial landmark $\overline{\mathbf{l}}\in{\mathbb{R}^{68\times 3}}$, the goal of Landmark LLE is to compute the reconstructed facial landmark $\overline{\mathbf{l}}^\prime\in{\mathbb{R}^{68\times 3}}$on each dimension. As is illustrated in Figure \ref{fig: overall}(c), we first find the $K$ nearest landmark of the input landmark $\overline{\mathbf{l}}$ in the GT landmark database $\mathcal{D}\in \mathbb{R}^{N\times 68\times 3}$ by computing the Euclidean distance, where $N$ is the number of landmark data points used to train the NeRF-based renderer.  After obtaining the $K$ nearest neighbors $\{\mathbf{l}_1,...,\mathbf{l}_K \}\subseteq \mathcal{D}$, we then seek a linear combination of these neighbors to reconstruct $\overline{\mathbf{l}}^\prime$ by minimizing the reconstruction error $||\overline{\mathbf{l}}-\overline{\mathbf{l}}^\prime||_2^2$, which could be formulated as the following least-squared optimization problem:
\begin{equation}
	\min ||\overline{\mathbf{l}}-\Sigma_{k=1}^K w_k \cdot \mathbf{l}_k||_2^2, ~~~~s.t. \Sigma_{k=1}^{K} w_k=1,
	\label{eq:least_square}
\end{equation}
where $w_k$ is the barycentric weight of the $k$-th nearest landmark $\mathbf{l}_k$. The optimial weights $\mathbf{w}^*=\{w_1^*,...,w_K^*\}\in \mathbb{R}^{K}$ can be obtained by solving Equation \ref{eq:least_square}. The hyper-parameter $K$ is chosen as 20 in our experiment via grid searching. Then we could compute the reconstructed facial landmark $\overline{\mathbf{l}}^\prime= \Sigma_{k=1}^K w_k^* \cdot \mathbf{l}_k$. Ideally, the reconstructed landmark $\overline{\mathbf{l}}^{\prime}$ can be regarded as an in-domain landmark data point that also possesses the semantic facial motion (such as eye blinking, laughing) in the input landmark $\overline{\mathbf{l}}$, though with some information loss. In practice, during inference, we use a linear combination of originally predicted landmark $\overline{\mathbf{l}}$ and reconstructed landmark $\overline{\mathbf{l}}^\prime$ as the final motion representation of the NeRF-based renderer: 
\begin{equation}
	\overline{\mathbf{l}}^{\prime\prime} = \overline{\mathbf{l}}^{\prime}\cdot \alpha + \overline{\mathbf{l}}\cdot(1-\alpha) = (\Sigma_{k=1}^K w_k^* \cdot \mathbf{l}_k)\cdot \alpha + \overline{\mathbf{l}}\cdot(1-\alpha),
	\label{eq:lle}
\end{equation}
where $\alpha\in [0,1]$ is a temperature hyper-parameter that tunes the trade-off between image quality and facial motion expressiveness: Intuitively, a large $\alpha$ would drag the $l$ closer to the GT dataset distribution and hence lead to better image quality and less rendering bad cases, while a smaller $\alpha$ would keep more details in the originally predicted landmark, which denotes more diverse and expressive facial motion. Our experiment results in Table \ref{tab:ablation_lle} show that the proposed method helps improve our system's robustness to predicted OOD landmarks. We provide a pseudo code of Landmark LLE in Appendix \ref{app:lle}.

\subsection{Instant Motion-to-Video Rendering}

In the previous sections, we obtain an expressive, time-consistent, and robust audio-to-motion mapping through the \textit{pitch-aware audio-to-motion} module and \textit{landmark LLE} postprocessing method. Next, we design an \textit{instant motion-to-video} module to efficiently render video frames conditioned on the predicted 3D landmarks, which is shown in Figure \ref{fig: overall}(d). 

\noindent \textbf{Grid-based NeRF Renderer} 
Recent progress in grid-based NeRF \cite{instant-ngp} proposes to encode 3D spatial information with a learnable feature grid. Compared with vanilla NeRF which obtains the spatial features with dense MLP forwarding, this new paradigm could directly query the features in the continuous 3D space via linear interpolation in the discrete feature grid, which is more efficient in both the training and inference stage. Following \cite{instant-ngp}, we utilize a learnable 3D grid to encode the queried position. Besides, an occupancy grid, which is a 3D grid that records the density value $\sigma$ estimated by NeRF, is maintained during training to prune the ray marching path. Since the utilization of a 3D feature grid eases the burden of NeRF to model the continuous spatial space, a lightweight grid-based NeRF could achieve comparable rendering quality with a deeper vanilla NeRF.

\noindent \textbf{Hyper-Space Landmark Encoder} 
To utilize the facial landmark to morph the human head in NeRF, some works \cite{park2021nerfies} adopt a conditional deformation field to warp the spatial points in the canonical space, which cannot model the non-rigid transform of the human head; GeneFace adopts a modulation-based method that directly concatenates the facial landmark and spatial features as the input to learn a landmark-conditioned NeRF. However, this method requires deep MLP forwarding (along with the input spatial feature)  to accurately morph the head geometry, which is computationally expensive and no longer feasible in a shallow grid-based NeRF. The motivation is to keep the accuracy of the input landmark to morph the head geometry while using a lightweight structure. 

Inspired by Hyper-NeRF \cite{park2021hypernerf}, as shown in Figure \ref{fig: overall}(d), we project the input facial landmark into an N-dimensional ambient coordinate conditioned on the grid-based spatial features, which allows an efficient fusion of spatial information and landmark condition. Once the ambient coordinate is obtained, instead of querying the landmark features with a dense MLP, we use an extra N-D learnable grid to improve efficiency. We empirically set $N=3$ via grid search, as a trade-off of performance and efficiency. The queried spatial features and landmark features are concatenated and fed into the NeRF backbone (a shallow MLP) to generate the density and color. Specifically, the implicit function can be formulated as:
\begin{equation}
	F:(f_x, f_l, d)\rightarrow c, \sigma
\end{equation} 
where $f_x$ and $f_l$ are the spatial/landmark feature queried from the grid, respectively. $d$ is the view direction. Then we could obtain a rendered image following the volume rendering technique illustrated in Equation \ref{eq:render_image} and train the renderer with Equation \ref{eq:train_nerf}.
	\section{Experiments}

\subsection{Experimental Setup} 

\noindent \textbf{Data Preparation and Preprocessing}
To learn a generalized audio-to-motion module, we use a clean subset of LRS3-TED \cite{afouras2018lrs3} to provide 190 hours of high-quality audio-motion pairs. To learn NeRF-based person-specific renderers, we adopt the dataset collected by \cite{LSP} and \cite{guo2021ad}, which consist of 5 videos of an average length of 6,000 frames in 25 FPS. During the data preprocessing phase, HuBERT \cite{hsu2021hubert} features and pitch contours are extracted from the audio track; head pose and 3D landmark are extracted from the video frames following \cite{ye2023geneface}. To train the NeRF, the target person videos are cropped into 512x512 resolution and each frame is processed with the help of an automatic parsing method \cite{lee2020maskgan} for segmenting the head and torso part and extracting a clean background.

\noindent \textbf{Compared Baselines} We compare our GeneFace++ with several remarkable works: 1)  Wav2Lip \cite{wav2lip}, which pretrains a sync-expert to improve the lip-synchronization performance; 2) MakeItTalk \cite{zhou2020makelttalk}, which also utilizes 3D landmark as the action representation; 3) LiveSpeechPortriat (LSP) \cite{LSP}, which achieves photorealistic results at over 30 FPS; 4) AD-NeRF \cite{guo2021ad}, which first utilize NeRF to achieve talking head generation. 5) RAD-NeRF \cite{tang2022real}, which adopts grid-based encoders in AD-NeRF to achieve real-time inference. 6) GeneFace \cite{ye2023geneface}, which achieves accurate lip-synchronization to OOD audios in NeRF-based talking face generation. 

\noindent \textbf{Model Configurations} 
GeneFace++ consists of a \textit{pitch-aware audio-to-motion} and an \textit{instant motion-to-video} module. The pitch-aware audio-to-motion module follows the major settings from GeneFace, with an additional pitch encoder. 
%In the pitch encoder, the number of bins is 300 in the discretization step, and the hidden size of pitch embedding and linear layer is 256. 
In the instant motion-to-video module, the network consists of two 3D learnable grid encoders to store the spatial and landmark information, respectively. The spatial and landmark information is then concatenated and fed into a shallow MLP backbone.
%, which consists of 5 fully connected layers to produce the density and color. 
For a fair comparison, the hyper-parameters of all NeRF-based baselines are adjusted to be coherent with our model. We provide detailed hyper-parameters of GeneFace++ in Appendix \ref{app:hparams}.

\noindent \textbf{Training Details} We train the GeneFace++ on 1 NVIDIA RTX 3090 GPU.  For VAE and Postnet in the pitch-aware audio-to-motion module, it takes about 40k and 10k steps to converge (about 12 hours). For the instant motion-to-video renderer, we train each model for 400k iterations (200k for the head and 200k for the torso, respectively), which takes about 10 hours.

\subsection{Quantitative Evaluation}
\noindent \textbf{Evaluation Metrics} We employ the FID score \cite{heusel2017fid} to measure image quality. We utilize the landmark distance (LMD)\cite{LMD} and syncnet confidence score (Sync score) \cite{wav2lip}  to evaluate audio-lip synchronization. To evaluate the inference speed, we implement a frame-per-second (FPS) assessment on a single NVIDIA 3090Ti GPU.

\begin{table}[t!]
	\centering
	%\small
	%\setlength{\tabcolsep}{5pt}
	\begin{tabular}{lccccc}
		\toprule
		
		\text{Method} & LMD$\downarrow$ & $\text{Sync}\uparrow$ &  $\text{PSNR}\uparrow$ & $\text{FID}\downarrow$  &   $\text{FPS}\uparrow$\\
		\midrule
		GT                   & 0.000  & 6.735& 0.00  & 0.000 & N/A   \\ 
		\midrule
		Wav2Lip         & 3.902 &  \textbf{7.618} & 29.16 & 71.963 & 11.95 \\ 
		MakeItTalk    & 4.838 &  4.383 & 27.94 & 50.087 & 22.03 \\ 
		LSP                 & 4.431 & 5.094  & 29.58 & 33.993&  25.35\\ 
		\midrule
		AD-NeRF       & 4.038 & 4.685 & 30.89 & 33.340 &  0.069 \\ 
		RAD-NeRF     & 3.984  & 4.886 & 31.02 & 33.029 & \textbf{28.37}\\ 
		GeneFace      & 3.827  & 5.596  & 31.08 & 29.677 &  0.064\\ 
		\midrule 
		GeneFace++ & \textbf{3.776}&  \textbf{6.114} & \textbf{31.22} &\textbf{29.147}&  23.55 \\ 
		\bottomrule
	\end{tabular}
	\caption{Quantitative evaluation with different methods. Best results are in \textbf{bold}.}
	\label{tab:main_results}
\end{table}

\noindent \textbf{Evaluation Results}
The results are shown in Table \ref{tab:main_results}. We have the following observations. (1) GeneFace++ achieves good \textit{audio-lip synchronization}, as it outperforms other baselines in terms of LMD and Sync score. \footnote{An exception is Wav2Lip in terms of Sync Socre, which is possibly due to the fact that Wav2Lip is jointly trained with SyncNet. As a result, it obtains a high sync score that is even higher than the ground truth video.}  (2) GeneFace++ achieves high \textit{video quality}, as it has the best performance in terms of PSNR and FID. (3) As for the \textit{system efficiency}, GeneFace++ could render talking face video in nearly real-time (about 23.55 FPS), which is a large acceleration from the GeneFace of 0.064 FPS. Both RAD-NeRF and GeneFace++ adopt the acceleration techniques from Instant-NGP to design the renderer, but RAD-NeRF achieves higher FPS with the comparable model scale of the renderer. This is due to the fact that GeneFace++ needs to execute an additional audio-to-motion and landmark LLE process to obtain lip-synchronized facial landmarks as the condition of the NeRF-based renderer. Although the audio-to-motion module and landmark LLE process are computationally cheap compared with the neural rendering process, it still brings slight latency in the overall system, which can be regarded as the sacrifices made to achieve a more accurate audio-lip synchronization.

\subsection{Qualitative Evaluation}

\begin{figure}
	\centering
	\includegraphics[width=0.8\textwidth]{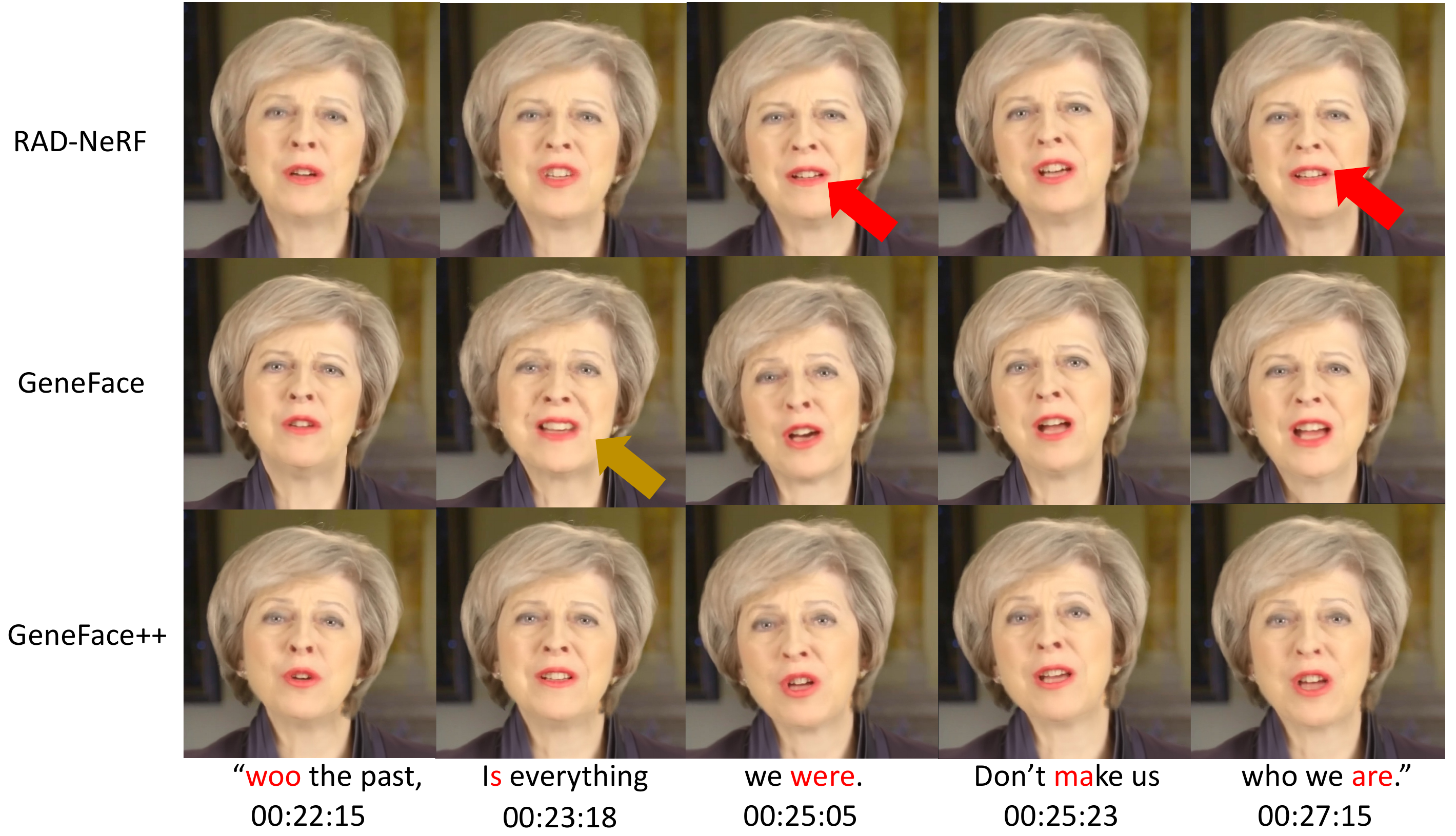}
	\caption{The comparison of generated key frame results. We show the speaking word and time step in the demo video. We mark the un-sync and bad rendering quality results with the red and brown arrows, respectively. \textbf{Please zoom in for better visualization}.}
	\label{fig:vis}
\end{figure}

To make a qualitative comparison of each method, we provide a demo video \footnote{The video URL is \url{https://genefaceplusplus.github.io/GeneFace++/dream_it_possible.mp4}} in which each method is driven by a three-minute-long English song as a hard case. We recommend that readers watch this video for a better comparison. We also show the keyframes of this demo video in Figure \ref{fig:vis}. Due to space limitations, we only compare our GeneFace++ with the two most competitive baselines (GeneFace and RAD-NeRF) here and provide a comparison with all baselines in Appendix \ref{app:qualitative}. We observe that GeneFace++ manages to handle several problems in previous NeRF-based methods: (1) while RAD-NeRF produces unsynchronized lip motion (red arrows in Figure \ref{fig:vis}) due to the weak generalizability to OOD audios, GeneFace++ achieves accurate lip motion. (2) GeneFace++ handles the occasionally occurred rendering bad cases in GeneFsce (brown arrows in Figure.\ref{fig:vis}), which are caused by OOD landmarks. (3) We also show that GeneFace++ improves the time consistency in the demo video.

\begin{table*}[]
	\centering
	\small
	\setlength{\tabcolsep}{3pt}
	\begin{tabular}{@{}l|ccccccc@{}}
		\toprule
		Methods      &   Wav2Lip & MakeItTalk &  LSP & AD-NeRF  &  RAD-NeRF  & GeneFace &\textbf{GeneFace++} \\ \midrule
		
		Audio-Lip Sync&     3.67$\pm$0.25   &    3.12$\pm$0.16       &   3.42$\pm$0.21 &  2.97$\pm$0.18   &  3.05$\pm$0.26  & 3.62$\pm$0.18 & \textbf{3.82$\pm$0.18 }\\
		
		Image Quality  &   2.84$\pm$0.25  &    2.67$\pm$0.32     &     3.55$\pm$0.23  &  3.31$\pm$0.20       &  3.43$\pm$0.24 & 3.39$\pm$0.21  &\textbf{ 3.58$\pm$0.18}  \\ 
		
		Video Realness       &     3.07$\pm$0.27 &    2.39$\pm$0.30     &   3.34$\pm$0.28     &      3.15$\pm$0.23    & 3.31$\pm$0.24& 3.39$\pm$0.22  & \textbf{3.54$\pm$0.26} \\  \bottomrule
	\end{tabular}
	\caption{User study with different methods. The error bars are 95\% confidence interval. }
	\label{tab:user_study}
\end{table*}

\paragraph{User Study}
We conduct user studies to test the quality of audio-driven portraits. Specifically, we sample 5 audio clips from English, French, and Italian (including a three-minute-long English song as a hard case) for all methods to generate the videos, and then involve 20 attendees for user studies. We adopt the Mean Opinion score (MOS) rating protocol for evaluation, which is scaled from 1 to 5. The attendees are required to rate the videos based on three aspects: (1) \textit{audio-lip synchronization}; (2) \textit{image quality}; (3) \textit{video realness}, which mainly measures the time consistency and 3D realness.

We compute the average score for each method, and the results are shown in Table \ref{tab:user_study}. We have the following observations: (1) Our GeneFace++ achieves the best \textit{audio-lip synchronization}, which shows the performance of our proposed pitch-aware audio-to-motion module. (2) As for the \textit{image quality}, with the delicately designed instant motion-to-video module, we found GeneFace++ achieves better image quality than previous person-specific NeRF-based (AD-NeRF, RAD-NeRF, and GeneFace) and GAN-based (LSP) renderers. We also observe that the one-shot talking face generation methods (Wav2Lip and MakeItTalk) obtains low scores in terms of image quality. (3) GeneFace also ahieves the best \textit{video realness } among the tested methods. We attribute the pitch-aware audio-to-motion module and landmark LLE method to good video realness, which could predict more accurate and time-consistent landmarks and automatically refine the outliers.

\subsection{Ablation Studies}
In this section, we perform ablation studies to prove the necessity of each component in GeneFace++.

\begin{table}
	\centering
	%\small
	%\setlength{\tabcolsep}{2pt}
	\begin{tabular}{lcccc}
		\toprule
		
		\text{Setting} &  $\text{L2 Error}\downarrow$ & \text{TE} & LMD$\downarrow$ & $\text{Sync}\uparrow$\\
		
		\midrule
		GeneFace++ & \textbf{0.0108} & \textbf{0.0031} & \textbf{3.776} & \textbf{6.114}  \\ 
		\midrule
		w/o P-VAE & 0.0133 & 0.0042& 3.783 & 5.975  \\ 
		w/o P-Postnet & 0.0192 & 0.0047& 3.810 & 5.732   \\ 
		GeneFace & 0.0237 & 0.0058& 3.827 & 5.596   \\ 
		\bottomrule
	\end{tabular}
	\caption{Ablation study results on pitch-aware audio-to-motion module. Temporal error is the L2 error on the velocity of the landmark sequence. }
	\label{tab:ablation_audio2motion}
\end{table}

\paragraph{Pitch-Aware Audio-to-Motion}
We test three settings on the audio-to-motion module: (1) removing pitch information in the VAE (w/o P-VAE); (2) removing pitch information in the Postnet (w/o P-Postnet); and (3) removing pitch information in VAE and Postnet, which is equivalent to the vanilla audio-to-motion module in GeneFace. We use the L2 error of the predicted landmark sequence to measure the lip accuracy and adopt a temporal error (which is the L2 error of the velocity of the landmark sequence) to measure the temporal consistency. We abbreviate temporal error as TE. The LMD and Sync score of the downstream rendered frames is also measured. The results are shown in Table \ref{tab:ablation_audio2motion}. We observe that removing pitch information leads to a significant degradation in lip accuracy and time consistency.

\begin{table}
	\centering
	%\small
	%\setlength{\tabcolsep}{3pt}
	\begin{tabular}{lccc}
		\toprule
		
		\text{Setting} &  $\text{LMD}\downarrow$ &  $\text{Sync}\uparrow$ &  $\text{BadCase}(\%)\downarrow$ \\
		
		\midrule
		$\alpha=0.0$ & \textbf{3.708} & \textbf{6.153} &  0.513 \\ 
		$\alpha=0.5$ & 3.776 & 6.114 &  0.164\\ 
		$\alpha=1.0$ & 3.825 & 6.039 & \textbf{ 0.102 }\\ 
		\bottomrule
	\end{tabular}
	\caption{Ablation study results on landmark LLE method. $\alpha$ is the hyper-parameter defined in Equation \ref{eq:lle}. $\text{BadCase\%}$ denotes the percentage of bad cases in the generated frames.}
	\label{tab:ablation_lle}
\end{table}

\paragraph{Landmark LLE}
We test three settings of Landmark LLE by tuning the hyper-parameter $\alpha$ defined in Equation \ref{eq:lle} from $\{0, 0.5, 1.0\}$. Note that $\alpha=0$ equals to not using Landmark LLE to post-process the predicted landmark and $\alpha=1.0$ means we use the LLE-reconstructed landmark $\mathbf{l}^{\prime}$ as the final input of the renderer. We are interested in two questions: (1) whether the additional postprocessing process would degrade the lip synchronization of the predicted landmark; (2) whether the proposed Landmark LLE method could effectively handle the bad cases caused by OOD landmarks. Therefore, we use LMD and Sync score to reflect the lip synchronization and compute the percentage of the bad case caused by OOD landmark in the generated videos, represented as $\text{BadCase}\%$. The results are shown in Table \ref{tab:ablation_lle}. We observe that adopting Landmark LLE significantly decreases the bad case rate, while slightly degrading the lip synchronization. As a trade-off, we use $\alpha=0.5$ in our experiments.

To further investigate the efficacy of Landmark LLE, we utilize T-SNE to visualize the landmarks at different stages in Appendix \ref{app:tsne}. We can see that Landmark LLE manages to drag outliers into the target person domain, hence improving the stability of the rendering results.

\paragraph{Instant-Motion-to-Video}

\begin{table}
	\centering
	\begin{tabular}{lccc}
		\toprule
		
		\text{Setting} &  $\text{PSNR(Face)}\uparrow$ &  $\text{PSNR(Lip)}\uparrow$ &  $\text{FPS}\uparrow$ \\
		
		\midrule
		dim=2 & 31.13 & 29.87 & \textbf{26.91} \\ 
		dim=3 &\textbf{ 31.22} & 30.19 &  23.55 \\ 
		dim=4 & 31.20 & \textbf{30.21} &  18.31   \\ 
		\bottomrule
	\end{tabular}
	\caption{Ablation study results on number of hyper coordinate dimensions of landmark.}
	\label{tab:ablation_hyper}
\end{table}

We ablate the number of hyper coordinate dimensions of the landmark encoder in the instant motion-to-video module. As a small coordinate dimension may limit the model's capacity and harms the rendering quality while a large coordinate requires more computation costs, we are interested in a suitable hyper coordinate dimension to efficiently model the landmark information. To this end, we test the number of dimensions within $\{2, 3, 4\}$. We take PSNR to measure the rendering quality and report the FPS to compare the inference speed. The results are shown in Table \ref{tab:ablation_hyper}. An interesting finding is that increasing the hyper coordinate dimension from $2$ to $3$ significantly improves the rendering quality on the Lip area. We suggest it is because the lip is a high-frequency dynamic part in a talking human head and could benefit from a larger model capacity. Therefore, we use a 3D hyper coordinate as a trade-off between rendering quality and inference speed.

	\section{Conclusion}

In this paper, we propose GeneFace++, which aims to achieve three goals in modern talking face generation: \textit{generalized audio-lip synchronization}, \textit{good video quality}, and \textit{high system efficiency}. GeneFace++ takes a big step forward from GeneFace. A pitch-aware audio-to-motion module is proposed to predict the facial landmark of generalized lip synchronization and high time consistency. A Landmark LLE method is introduced to automatically refine the predicted landmark and avoid the potential bad cases of rendering. A delicate instant motion-to-video renderer is designed to generate high-quality video efficiently. Extensive experiments show that our method achieves the three goals of the modern talking face system and outperforms existing methods. Due to space limitations, we discuss the limitations and future works in Appendix \ref{app:limitation}.

%\section*{Acknowledgements}
%This work was supported in part by the Zhejiang Natural Science Foundation LR19F020006 and National Key R\&D Program of China under Grant No.2020YFC0832505.

	%\newpage
	\bibliographystyle{plain}
	\bibliography{main}
	
	\appendix
	% \clearpage

\section{Details of Models}
\subsection{Detailed Network Structure}
\begin{figure*}[ht]
	\centering
	\includegraphics[width=1.0\textwidth]{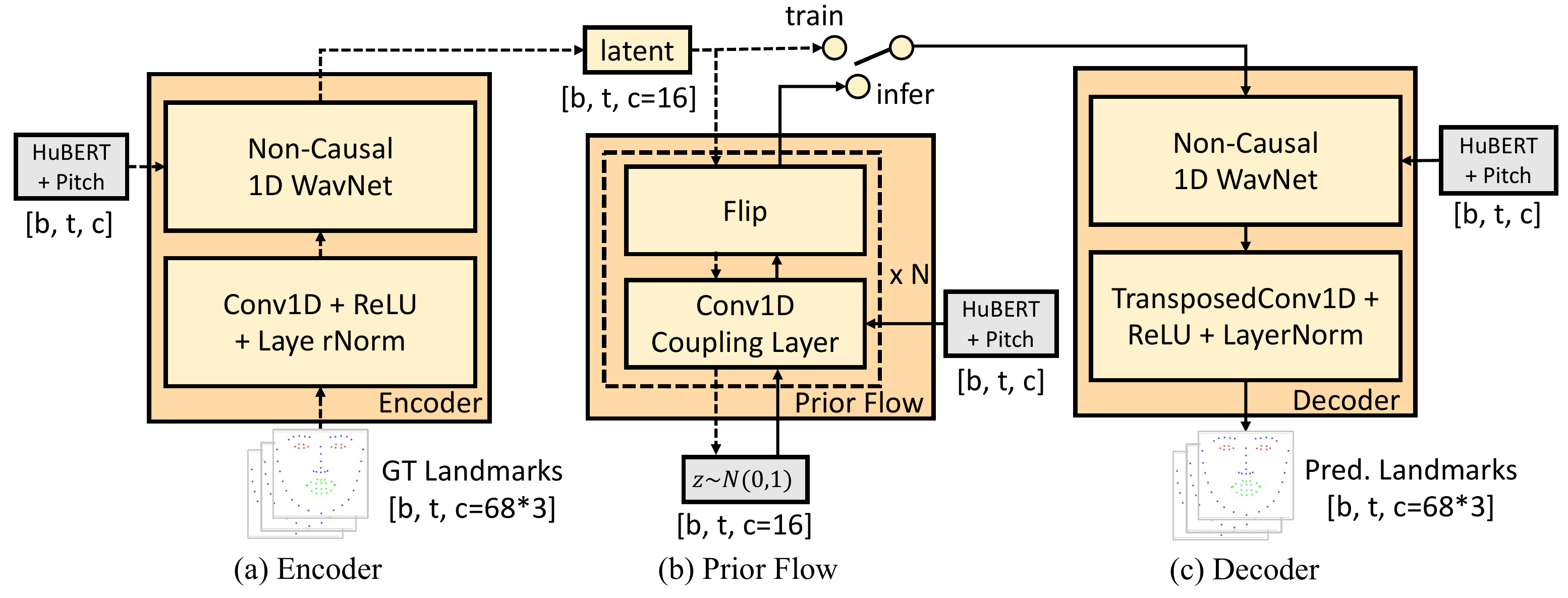}
	\caption{The detailed structure of VAE in the Pitch-Aware Audio-to-Motion module. In subfigure (a), (b), and (c), we show the structure of encoder, prior flow, and the decoder, respectively. Dotted arrows are only operated at the training phase. "Pred." is a shorthand of "predicted".
		% In subfigure (a), Enc. and Dec. denote Encoder and Decoder in VAE, respectively. Disc. means Discriminator. In subfigure (b), the thunder-like symbol represents the "stop gradient" operator. In subfigure(c), "Log-Discretize" denotes the operation that quantizes the log-scale continuous pitch value into discrete tokens.
	}
	\label{fig: vae}
\end{figure*}

\begin{figure*}[ht]
	\centering
	\includegraphics[width=0.9\textwidth]{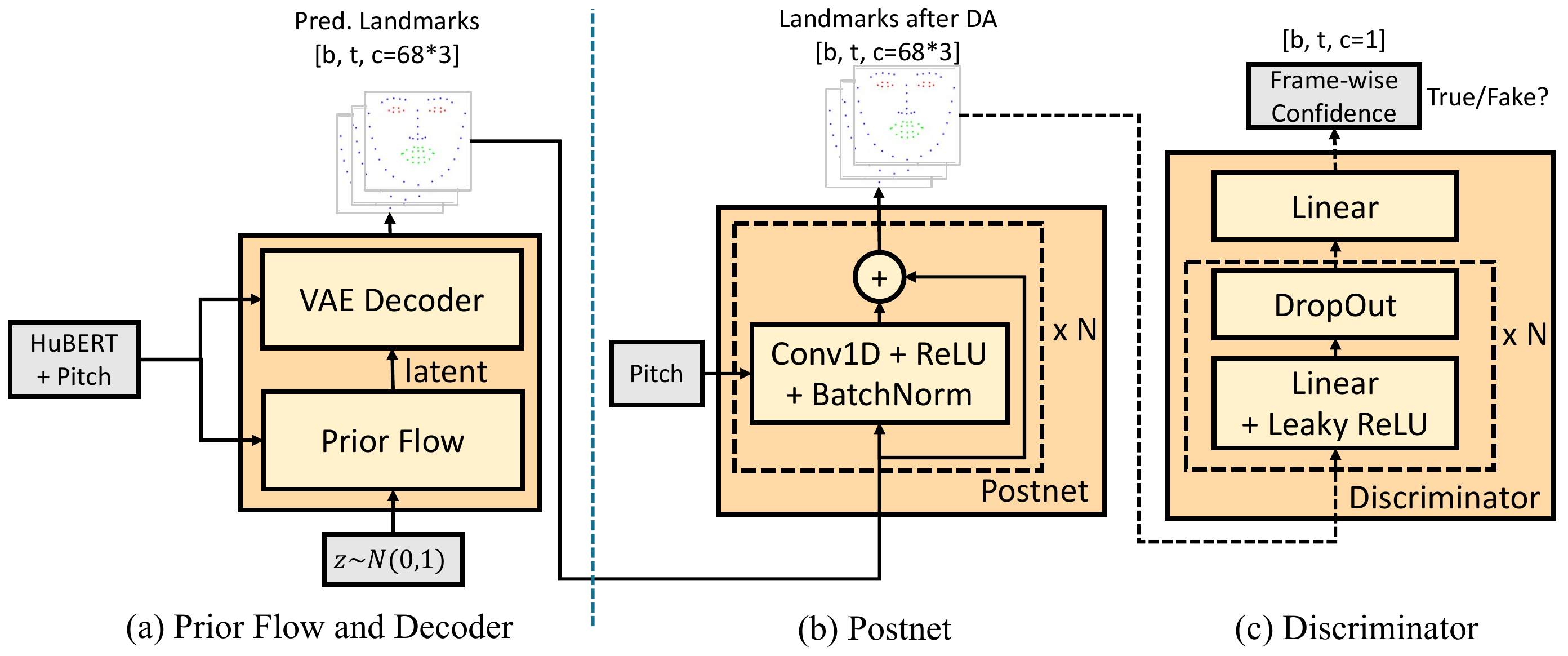}
	\caption{The detailed structure of Postnet and its discriminator in the Pitch-Aware Audio-to-Motion module. In subfigure (a), Prior Flow and Decoder are pretrained and fixed to predict raw landmark sequence. We separated them with a blue dotted line. In subfigure (b) and (c) we show the structure of Postnet and its discriminator, respectively. Dotted arrows are only operated at the training phase. "Pred." is a shorthand of "predicted".
		. 
		% In subfigure (a), Enc. and Dec. denote Encoder and Decoder in VAE, respectively. Disc. means Discriminator. In subfigure (b), the thunder-like symbol represents the "stop gradient" operator. In subfigure(c), "Log-Discretize" denotes the operation that quantizes the log-scale continuous pitch value into discrete tokens.
	}
	\label{fig: postnet}
\end{figure*}

We provide the detailed network structure of audio-to-motion VAE and Postnet in Figure \ref{fig: vae} and Figure \ref{fig: postnet}, respectively. As shown in \ref{fig: vae}(b), in practice, we additionally train a flow-based model to predict the latent code during inference, which is adhere to GeneFace.

\label{app:structure}

\subsection{Training Losses of Audio-to-Motion Module}
\label{app:loss_audio2motion}
\noindent \textbf{Pitch-Aware VAE}
We add a temporal term $l_{temp}$ in the training of VAE and Postnet, which is the L2 error of the velocity of the landmark sequence and could be represented as:
\begin{equation}
	l_{temp}(v, \hat{v}) = ||v-\hat{v}||_2^2
\end{equation}
where $v$ and $\hat{v}$ is the velocity of GT landmark $l$ and predicted landmark $\hat{l}$, respectively. Now that the temporal term is defined, the training loss of Pitch-Aware VAE could be formulated on the top of Equation \ref{eq:vae}:
\begin{equation}
	\mathcal{L}_{\text{P-VAE}}=\mathbb{E}[ ||l - \hat{l}||_2^2  + {KL}(z| \hat{z}) + l_{Sync}(a, \hat{l}) + l_{temp}(v,\hat{v})].
	\label{eq:vae_pitch}
\end{equation}

\noindent \textbf{Pitch-Aware Postnet}
As for the training of Pitch-Aware Postnet, the Postnet follows the Equation \ref{eq:pn} in GeneFace, and the discriminator is enhanced by introducing a gradient penalty term:
\begin{equation}
	l_{gp}(l) = \nabla_l Disc(l),
\end{equation}
where $l$ is the GT landmark. This term encourages the discriminator to give a similar confidence score to a similar input landmark, which successfully prevents overfitting in the adversarial training. Now that the gradient penalty term is obtained, we formulate the loss of discriminator as:

\begin{equation}
	\mathcal{L}_{\text{Disc}}= \mathbb{E}[l_{adv}(l, \hat{l}) + l_{gp}(l)],
\end{equation}
where $l_{adv}$ is the discriminator loss of LS-GAN.

\subsection{Pseudo Code of Landmark LLE Method}
\label{app:lle}
We provide the pseudo code of the proposed Landmark LLE method in Algorithm \ref{alg:lle}.
\begin{algorithm}[ht]
	\centering
	\caption{Landmark Locally Linear Embedding Method}\label{alg:lle}
	\begin{algorithmic}[1]
		\STATE \textbf{Input}: GT landmark set $\mathcal{D}$; Predicted landmark $\mathbf{l}$.
		\STATE \textbf{Output}: Projected landmark $\mathbf{l}^{\prime}$.
		\STATE Compute $K$ nearest neighbors of $\mathbf{l}$ in the GT landmark set $\mathbf{L}$.
		\STATE Solve the least squared problem defined in Equation \ref{eq:least_square}, obtain the optimal weights of $K$ neighbors $\mathbf{w}^*=\{w^*_1, ..., w^*_K\}$.
		\STATE Copute the projected landmark following Equation \ref{eq:lle}.
	\end{algorithmic}
\end{algorithm}

\section{Detailed Experimental Settings}
\subsection{Model Configurations}
\label{app:hparams}
We list the hyper-paramters	of GeneFace++ in Table \ref{tab:hparams}.

\begin{table*}[htbp]
	\centering
	\small
	\caption{Hyper-parameter list}
	\begin{tabular}{ccc}
		\hline
		\multicolumn{2}{c|}{Hyper-parameter} & Value \\
		\hline
		\multicolumn{1}{c|}{\multirow{5}[3]{*}{Pitch Encoder}} & \multicolumn{1}{l|}{ Number of Pitch Bins} & 300   \\
		\multicolumn{1}{c|}{} & \multicolumn{1}{l|}{Pitch Embedding Channel Size} & 64  \\
		\multicolumn{1}{c|}{} & \multicolumn{1}{l|}{Number of Conv1D Layers} & 2 \\
		\multicolumn{1}{c|}{} & \multicolumn{1}{l|}{Pitch Encoder Conv1D Kernel} & 3      \\
		\multicolumn{1}{c|}{} & \multicolumn{1}{l|}{Pitch Encoder Conv1D Channel Size} & 64    \\
		
		\hline
		\multicolumn{1}{c|}{\multirow{8}[5]{*}{Audio-to-Motion VAE}} & \multicolumn{1}{l|}{Encoder Layers} & 8    \\
		\multicolumn{1}{c|}{} & \multicolumn{1}{l|}{Decoder Layers} & 4       \\
		\multicolumn{1}{c|}{} & \multicolumn{1}{l|}{Encoder/Decoder Conv1D Kernel} & 5       \\
		\multicolumn{1}{c|}{} & \multicolumn{1}{l|}{Encoder/Decoder Conv1D Channel Size} & 192     \\
		\multicolumn{1}{c|}{} & \multicolumn{1}{l|}{Latent Size } & 16      \\
		\multicolumn{1}{c|}{} & \multicolumn{1}{l|}{Prior Flow Layers} & 4      \\
		\multicolumn{1}{c|}{} & \multicolumn{1}{l|}{Prior Flow Conv1D Kernel} & 3      \\
		\multicolumn{1}{c|}{} & \multicolumn{1}{l|}{Prior Flow Conv1D Channel Size} & 64    \\
		\multicolumn{1}{c|}{} & \multicolumn{1}{l|}{Sync-expert Layers} & 14    \\
		\multicolumn{1}{c|}{} & \multicolumn{1}{l|}{Sync-expert Channel Size} & 512    \\
		\hline
		\multicolumn{1}{c|}{\multirow{6}[4]{*}{Post-net and its DA-Discriminator}} & \multicolumn{1}{l|}{Post-net Layers} & 8   \\
		\multicolumn{1}{c|}{} & \multicolumn{1}{l|}{Post-net Conv1D Kernel} & 3  \\
		\multicolumn{1}{c|}{} & \multicolumn{1}{l|}{Post-net Conv1D Channel Size} & 256  \\
		\multicolumn{1}{c|}{} & \multicolumn{1}{l|}{Discrimnator Layers} & 5      \\
		\multicolumn{1}{c|}{} & \multicolumn{1}{l|}{Discrimnator Linear Hidden Size} & 256    \\
		\multicolumn{1}{c|}{} & \multicolumn{1}{l|}{Discrimnator Dropout Rate} & 0.25    \\
		\hline
		\multicolumn{1}{c|}{\multirow{2}[1]{*}{Landmark LLE}} & \multicolumn{1}{l|}{Number of Neighbors ($K$)} & 20   \\
		\multicolumn{1}{c|}{} & \multicolumn{1}{l|}{Weights of LLE results to construct the final landmark ($\alpha$)} & 0.5  \\
		\hline
		
		\multicolumn{1}{c|}{\multirow{8}[4]{*}{Instant Motion-to-Video Renderer}} & \multicolumn{1}{l|}{Head/Torso NeRF Layers} & 5   \\
		\multicolumn{1}{c|}{} & \multicolumn{1}{l|}{Head/Torso NeRF Hidden Size} & 64  \\
		\multicolumn{1}{c|}{} & \multicolumn{1}{l|}{Head NeRF Spatial Grid Dimension} & 3  \\
		\multicolumn{1}{c|}{} & \multicolumn{1}{l|}{Head NeRF Spatial Grid Channel Size per Resolution Revel} & 2  \\
		\multicolumn{1}{c|}{} & \multicolumn{1}{l|}{Head NeRF Spatial Grid Number of Resolution Revel} & 16  \\
		\multicolumn{1}{c|}{} & \multicolumn{1}{l|}{Landmark Encoder Layers} & 6      \\
		\multicolumn{1}{c|}{} & \multicolumn{1}{l|}{Landmark Encoder Hidden Size} & 64    \\
		\multicolumn{1}{c|}{} & \multicolumn{1}{l|}{Landmark Hyper Grid Dimension} & 3  \\
		\multicolumn{1}{c|}{} & \multicolumn{1}{l|}{Landmark Hyper Grid Channel Size} & 2  \\
		\hline
	\end{tabular}%
	\label{tab:hparams}%
\end{table*}%

\section{Additional Experiments} 
\subsection{Qualitative Results with All Baselines}
\label{app:qualitative}
\begin{figure*}[ht]
	\centering
	\includegraphics[width=0.55\textwidth]{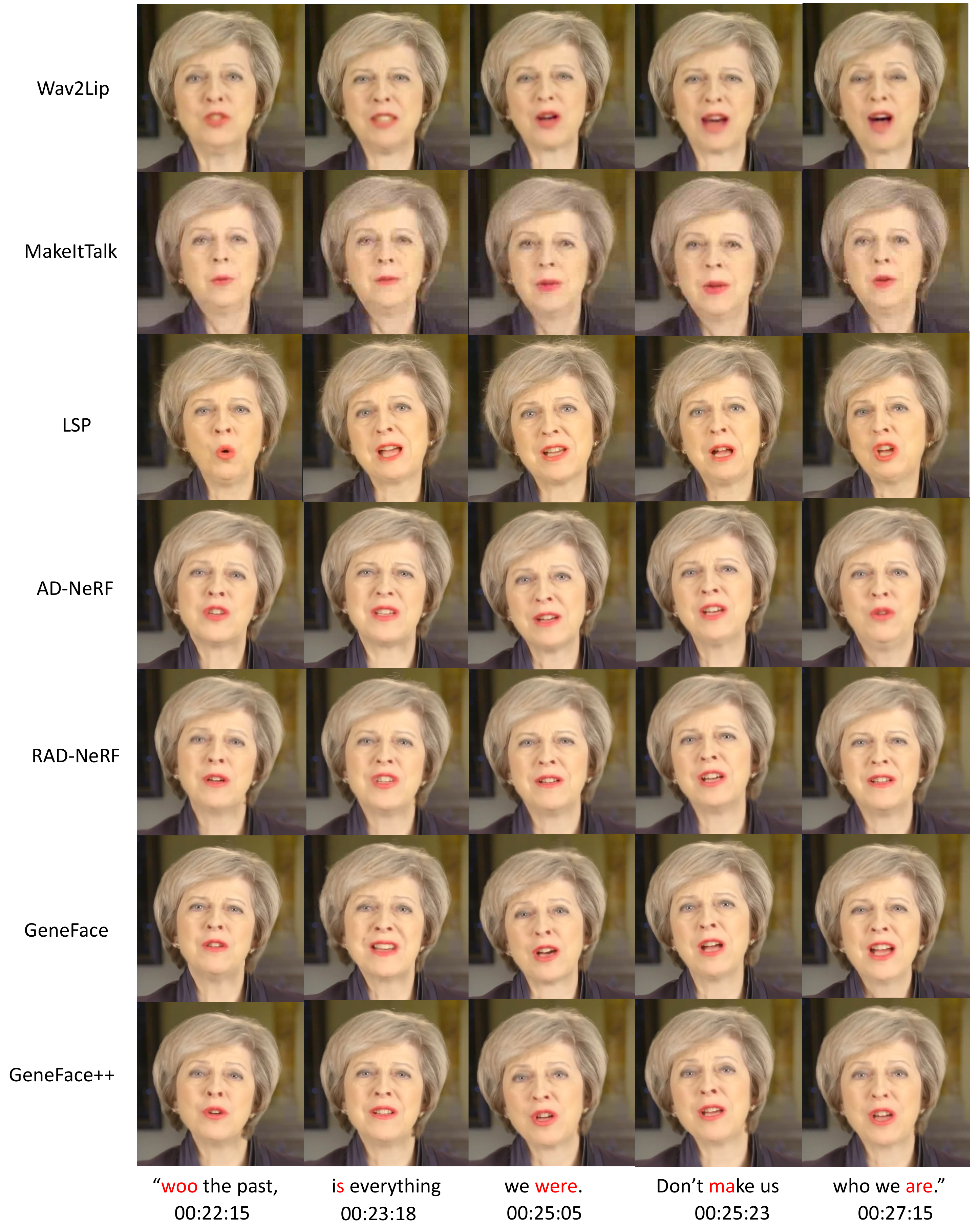}
	\caption{The comparison of generated key frame results. We show the speaking word and time step in the demo video.  \textbf{Please zoom in for better visualization}.
	}
	\label{fig:vis_full}
\end{figure*}

We show the rendered keyframes of different talking face methods in Figure \ref{fig:vis_full}. We recommend the readers watch the demo video for a better comparison. The video URL is \url{https://genefaceplusplus.github.io/GeneFace++/dream_it_possible.mp4}

\subsection{T-SNE Visualization of 3DMM Landmark}
\label{app:tsne}
\begin{figure*}[ht]
	\centering
	\includegraphics[width=0.65\textwidth]{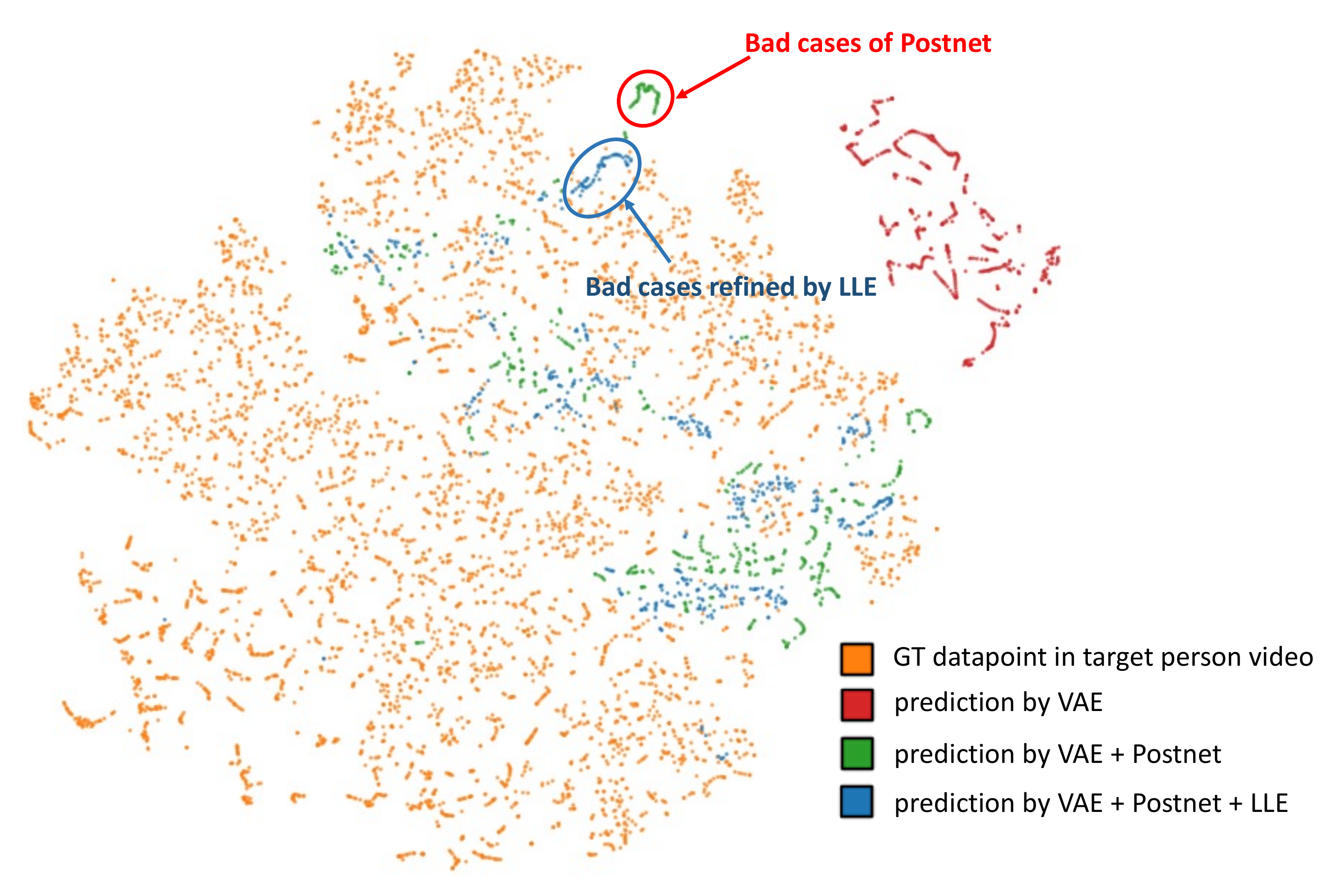}
	
	\caption{The T-SNE visualization of 3DMM landmarks in the different stages of GeneFace++.  }
	\label{fig:tsne}
\end{figure*}

To validate the effectiveness of the proposed Landmark LLE method, we adopt T-SNE to visualize the predicted landmark at different stages of GeneFace++. The result is shown in Figure \ref{fig:tsne}. The brown points denotes the ground truth landmarks in the target person video. Ideally the predicted landmarks should within this distribution so that the motion-to-video module could render images of good quality. The red points are landmarks predicted by the VAE, and the green points are the prediction refined by the Postnet. We can see that Postnet manages to project the marjority of predicted landmark into the target person domain, yet there still exist some outliers out of the distribution (marked with a red cycle) and will raise rendering bad cases. Then the blue points are the predicted landmark further postprocessed by the Landmark LLE method. We can see that LLE successfultly drag the outliers into the target person video, hence significantly improves the stability of the rendering result.

\section{Limitation and Future Works}
\label{app:limitation}
To validate the effectiveness of the proposed Landmark LLE method, we adopt T-SNE to visualize the predicted landmark at different stages of GeneFace++. The result is shown in Figure \ref{fig:tsne}. The brown points denote the ground truth landmarks in the target person's video. Ideally, the predicted landmarks should be within this distribution so that the motion-to-video module could render images of good quality. The red points are landmarks predicted by the VAE, and the green points are the prediction refined by the Postnet. We can see that Postnet manages to project the majority of predicted landmarks into the target person domain, yet there still exist some outliers out of the distribution (marked with a red cycle) and will raise rendering bad cases. Then the blue points are the predicted further post-processed by the Landmark LLE method. We can see that LLE successfully drags the outliers into the target person's video, hence significantly improving the stability of the rendering result.

	%%%%%%%%%%%%%%%%%%%%%%%%%%%%%%%%%%%%%%%%%%%%%%%%%%%%%%%%%%%%

\end{document}